\documentclass[Afour,sageh,times]{sagej}
\usepackage[utf8]{inputenc}

\usepackage[colorlinks,bookmarksopen,bookmarksnumbered,citecolor=red,urlcolor=red]{hyperref}

\newcommand\BibTeX{{\rmfamily B\kern-.05em \textsc{i\kern-.025em b}\kern-.08em
T\kern-.1667em\lower.7ex\hbox{E}\kern-.125emX}}

\def\volumeyear{2019}

\usepackage{caption}
\usepackage{subfigure}

\begin{document}

\runninghead{Schillaci et al.}

\title{Intrinsic Motivation and Episodic Memories for Robot Exploration of High-Dimensional Sensory Spaces}

\author{Guido Schillaci\affilnum{1,2}, Antonio Pico Villalpando\affilnum{3}, Verena V. Hafner\affilnum{3}, Peter Hanappe\affilnum{4}, David Colliaux\affilnum{4}, Timoth\'{e}e Wintz\affilnum{4}}

\affiliation{\affilnum{1} The BioRobotics Institute, Scuola Superiore Sant'Anna, Pisa, Italy\\
\affilnum{2} Department of Excellence in Robotics \& AI, Scuola Superiore Sant'Anna, Pisa, Italy\\
\affilnum{3} Adaptive Systems Group, Humboldt-Universit\"{a}t zu Berlin, Germany
\\
\affilnum{4} Sony Computer Science Laboratories, Paris, France}

\corrauth{Guido Schillaci, Dr. rer. nat.\\
Marie Skłodowska Curie Fellow\\
The BioRobotics Institute, Scuola Superiore Sant’Anna\\
Viale Rinaldo Piaggio, 34, 56025 - Pontedera (PI) - Italy}

\email{guido.schillaci@santannapisa.it}

\begin{abstract}
This work presents an architecture that generates curiosity-driven goal-directed exploration behaviours for an image sensor of a microfarming robot. A combination of deep neural networks for offline unsupervised learning of low-dimensional features from images, and of online learning of shallow neural networks representing the inverse and forward kinematics of the system have been used. The artificial curiosity system assigns interest values to a set of pre-defined goals, and drives the exploration towards those that are expected to maximise the learning progress. We propose the integration of an episodic memory in intrinsic motivation systems to face catastrophic forgetting issues, typically experienced when performing online updates of artificial neural networks. Our results show that adopting an episodic memory system not only prevents the computational models from quickly forgetting knowledge that has been previously acquired, but also provides new avenues for modulating the balance between plasticity and stability of the models. 
\end{abstract}


\maketitle

\section{Introduction}

Intrinsic motivation refers to the act of engaging in a pleasurable activity, where the satisfaction or reward is not coming from an external source, but from the activity itself. In psychology and robotics, this is connected to the cognitive phenomenon of curiosity, a form of intrinsic motivation that drives behaviours towards novel and surprising activity \citep{oudeyer2007intrinsic,oudeyer2016intrinsic}. 

Exploration and play seem to be partially driven by intrinsic motivation in infancy \citep{oudeyer2007intrinsic}. Visual exploration is likely to be one of the earliest behaviours influenced by this drive \citep{schlesinger2013investigating}, although studies in prenatal development suggest that other modalities, such as touch \citep{zoia2013development}, may be more predominant.

In developmental sciences, intrinsic motivation and curiosity are topics of great interest. 
It is through exploration and play - likely affected by these drives - that infants incrementally learn about their bodily capabilities and about how to interact with their surroundings \citep{baldassarre2013intrinsically}. 
In developmental robotics, these processes have been demonstrated to be promising tools for enabling learning, adaptivity and curiosity-driven behaviours in artificial agents (see for instance \citep{forestier2017intrinsically,colas2019curious,frank2014curiosity} and \citep{schillaci2016} for a review). 
Intrinsic motivation algorithms drive the actions of an artificial agent towards activities that are expected to maximise the information gain, and thus the learning progress, when this is integrated within a learning framework. These expectations are generated by predictive models, which represent the core components of such systems, allowing the agent to anticipate the sensory consequences of self-generated actions. Prediction errors can be computed as the discrepancy between the expected sensory inputs and the actual sensory observations. The system thus monitors the dynamics of such prediction errors over time, and selects those behaviours that are expected to produce big variations in the prediction errors - or, in other words, those activities that may be generating a high information gain.

Traditionally, algorithms implementing intrinsic motivation have been applied in the context of learning motor control \citep{baranes2013active,oudeyer2007intrinsic,baldassarre2013intrinsically}. This approach, combined with goal-directed exploration strategies \citep{rolf2012goal,Schmerling2015}, has shown to be highly efficient when learning controllers for high-dimensional actuators. 
However, in such studies, the sensory space is often over-simplified or low-dimensional. For instance, it is represented as the Cartesian coordinates of the end-effector of the robot - since the main focus is rather on learning controllers of complex, high-dimensional and redundant robot manipulators.

Very few studies addressed more realistic sensory spaces (e.g. full resolution visual inputs) in the literature on intrinsic motivation algorithms for robots. \citep{santucci2016grail} proposed an architecture for intrinsically motivated exploration in a humanoid robot, where the sensory space is represented by the visual input grabbed from the robot camera in a simulated environment. The authors studied how to represent goals corresponding to changes in the environment. The world consisted of a simulated environment with few rounded objects placed in front of the robot. Visual inputs and goals were encoded as binary images. 

Studies that specifically address the visual modality in exploratory behaviours can be found in the wide literature on active vision. Active vision is a robotic application in which the pose and the configuration of a visual sensor is determined for solving vision-based tasks, usually those that require obtaining multiple views of an object to be manipulated (see \citep{chen2011active} for a review).
Although interesting studies can be found in this literature, they mostly focus on the specific tasks to be tackled by the active visual perception and lack of generalisation to more complex and integrated sensorimotor representations. Moreover, they miss the explanatory potential that intrinsic motivation research has towards the functioning of higher cognitive processes (e.g. curiosity and learning).

There is however an important challenge in intrinsic motivation systems. Intrinsically motivated agents collect data and acquire skills in an incremental fashion through the online self-generation of training samples \citep{parisi2019continual}. Learning is tightly coupled with the behaviour of the agent. When "wrong" behaviours are generated in the initial phases of the learning sessions, the computational model may converge to sub-optimal solutions and local minima, thus preventing the system to generate further explorative behaviours. Adaptive systems should be escaping such situations, perhaps 

Strategies to avoid this issue imply a good balance between instrumental (goal-directed) actions and epistemic (novelty-seeking) actions, so that bad-bootstrapping of models can be avoided \citep{tschantz2019learning}. Different solutions have been proposed, including $\epsilon$-greedy strategies \citep{oudeyer2007intrinsic} (generating random actions or replaying previous experience to the system) or active inference approaches \citep{tschantz2019learning}.

Related to this issue is also the problem of finding an appropriate balance between plasticity and stability of the models, when training them in an online fashion \citep{mermillod2013stability}. Typically, when a model is being trained with new information, previously learned knowledge quickly becomes overwritten by the new one. This is a well-known issue in machine learning, especially in the context of artificial neural networks, named \textit{catastrophic forgetting} \citep{mcclelland1995there}.
In fact, the training on new samples may disrupt connection weights in the neural network that were encoding previous mappings \citep{Masse2018}. Different strategies have been proposed to overcome this problem. One of such approaches is known as \textit{memory consolidation} or \textit{system-level consolidation} \citep{mcclelland1995there}: an episodic memory system maintains a subset of previously experienced sensorimotor data and replays them, along with the new samples, to the networks during the training. 

The work presented here approaches the aforementioned challenges by combining online deep learning, intrinsic motivation and a memory consolidation system. To the best of our knowledge, this is one of the first studies in the literature combining these features.
In particular, we present an architecture that generates curiosity-driven goal-directed exploration behaviours in a robot, using computational models that can deal with high dimensional sensory inputs. The computational models are trained in an online fashion throughout the exploratory behaviours generated by the intrinsic motivation system.
Thanks to the adoption of an episodic memory, the system is less prone to catastrophic forgetting. 
A combination of deep and shallow neural networks has been used. In particular, deep convolutional neural networks are adopted for offline unsupervised learning of low-dimensional features from images. Online learning is, instead, performed on shallow neural networks encoding the internal models (i.e. inverse and forward kinematics) of the robot. The intrinsic motivation system relies on a set of pre-defined goals, and drives the exploration towards those goals which 
are expected to maximise the learning progress. We adopt an episodic memory system to face catastrophic forgetting issues experienced when performing online updates of the internal models of the robot.

We test the framework on a simulator of a microfarming robot. 
Microfarms present interesting challenges for robotics applications. These farms are characterised by small surfaces (0.01 to 5 ha) and typically grow a much larger variety of crops than conventional farms. A considerable amount of work is still carried out manually, as their limited size and their diversity prevents the usage of typical agricultural machines, such as tractors. Open and lightweight robots for microfarms may reduce manual labour and increase their productivity. However, the robots must be able to cope with many types of plants and dense plant populations.

A first step in the development of such advanced applications is to automatically construct a 3D representation of plants in outdoor settings. The 3D scanning procedure collects many images of the object of interest using either multiple fixed cameras or a single one that moves around the object along a pre-defined trajectory. The image set is converted into a point cloud (e.g. as in structure-from-motion algorithms \citep{schoenberger2016sfm,schoenberger2016mvs}) and then post-processed for estimating 3D models or for detecting plant organs.

Plants are, however, complex objects to reconstruct and many parts are hidden, for example, by leaves. If the camera movements are not sufficiently accurate to uncover such hidden spots, the reconstruction process may generate incomplete 3D models. The trajectory of the camera movements must therefore adapt to the plant of interest.
Intelligent and adaptive behaviours - like those generated by the proposed framework - could maximise the information captured by the robot camera. 

The rest of this paper is structured as follows. First, we introduce the robotic platform (Section \ref{sec:robot}) and the simulator (Section \ref{sec:simulator}) that have been used in this study. In Section \ref{sec:architecture}, we describe the learning architecture, including the computational models, the goal selection strategy, the image encoding and the main learning algorithm.
Sections \ref{sec:experiments} and \ref{sec:results} describe the experiments and the results, respectively. 
In particular, we show the performance of the system under different configurations, giving a particular emphasis on the episodic memory system. We draw our conclusions in section \ref{sec:conclusions}, where we present the plan for future work.

\section{Experimental setup}
\subsection{Robotic platform}
\label{sec:robot}
\begin{figure}
	\centering
	\includegraphics[width=8cm]{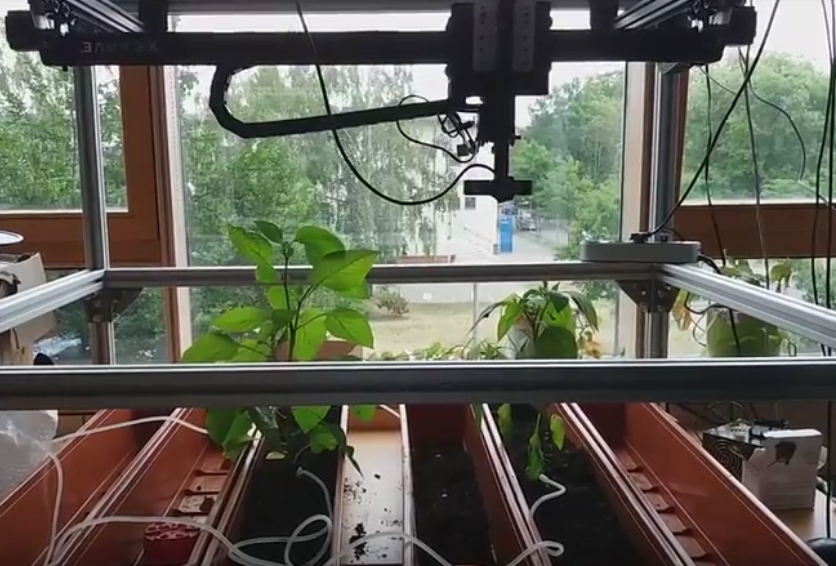}
	\caption{The LettuceThink micro-farming robot.}
	\label{fig:lettucethink}
\end{figure}

This work presents an architecture that generates curiosity-driven goal-directed exploration behaviours for an image sensor of a microfarming robot. The studies reported here have been carried out on a simulated version (described in Section \ref{sec:simulator}) of the LettuceThink micro-farming robot (Figure \ref{fig:lettucethink}) developed by Sony Computer Science Laboratories. The platform consists of an aluminium frame with an X-Carve CNC machine mounted on it. The CNC machine is used to provide 3-axes movements to a depth camera (Sony DepthSense) mounted at the tip of the vertical z-axis (hereafter, the \textit{end-effector camera}). In the experiments presented in this paper, the end-effector camera is facing top-down and only two motors are used (x and y). This configuration is not intended to be used for 3D reconstruction of plants, which requires more degrees of freedom for the camera movements. 

A Raspberry Pi embedded computer connects to the X-Carve controller over a serial connection and to the end-effector depth camera over USB. 
The X-Carve controller consists of an Arduino board running the Grbl firmware, an open source software that interprets G-code instructions and controls the stepper motors of the CNC machine\footnote{For further informations, please refer to https://github.com/grbl/grbl/wiki}.
The stepper motors are not equipped with rotary encoders but their status are obtained through a Grbl WPos (working position) request that returns the offset of each motor from the initial position by counting the steps performed from the beginning of the operation. The Raspberry Pi uses the ROS (Robot Operating System) middleware to let external computers control the robot hardware and to synchronise the motor and image data.

\subsection{Robot simulator}
\label{sec:simulator}
A simulator of the LettuceThink robot has been developed to ease the testing of different configurations of the learning system. The simulator generates sensorimotor data from requested trajectories of the end-effector camera. Knowing the initial position of the CNC machine and the target position, the simulator linearly interpolates the trajectory and returns the intermediate positions of the camera together with the images captured from each specific position. The sensorimotor data returned by the simulator have been pre-recorded by performing a full scan of the (x,y) plane of the CNC machine using a resolution of 5mm. This resulted in 24,964 images, each mapped to an (x,y) position of the CNC machine.

The simulator is freely available as as a python script that generates sensorimotor data\footnote{The script will be available here: \url{https://github.com/guidoschillaci/sonylettucethink_dataset}} from the recorded dataset\footnote{The repository containing the images will be made freely downloadable in a ZENODO.}. All the software developed for this work is freely available in an online repository\footnote{The latest version can be downloaded here: \url{https://github.com/guidoschillaci/goal_babbling_cae_episodic_memory}.}.

\section{Learning architecture}
\label{sec:architecture}
The architecture implementing the curiosity-driven goal-directed exploration behaviours combines one deep and two shallow neural networks, as well as an episodic memory system.
A deep convolutional neural network is trained offline using unsupervised methods to encode images using low-dimensional features. The shallow neural networks learn online how to represent the inverse and forward kinematics of the robotics system. In addition, an artificial curiosity system assigns interest values to a set of pre-defined goals and drives the exploration towards goals that are expected to maximise the learning progress. An \textit{episodic memory system} is included to handle catastrophic forgetting issues.

The learning architecture controls the movements of the end-effector camera of the LettuceThink platform using an exploration behaviour driven by artificial curiosity. 
Differently to adopting pre-defined trajectories, the behaviours generated by this system aim at autonomously maximising the information obtained with the image sensors. 
In particular, the aim of the learning architecture is to direct the movements of the camera towards positions that produce informative views of objects of interest, and in parallel to update the internal models with the sensorimotor data generated by the latest movements.

\begin{figure*}
	\centering
	\includegraphics[width=10.5cm]{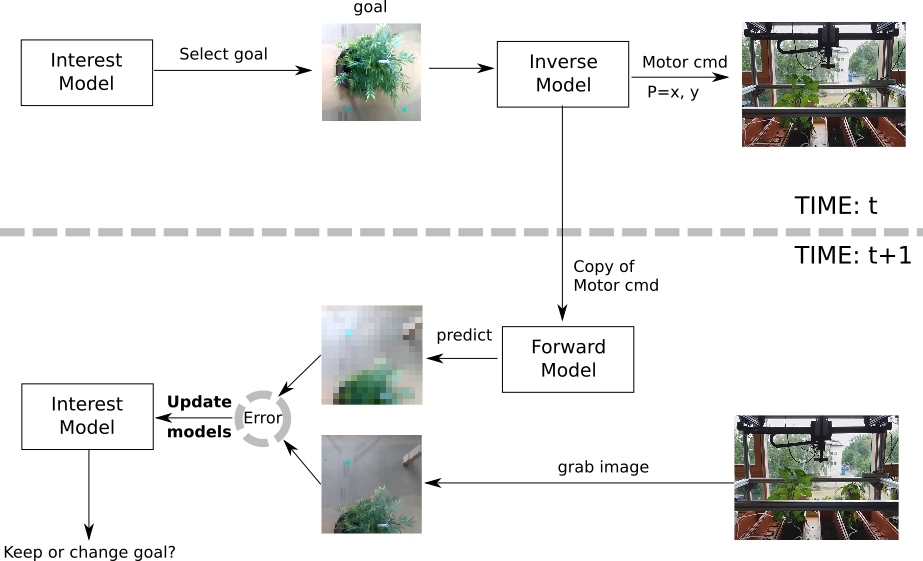}
	\caption{The architecture that generates goal-directed movements driven by artificial curiosity. In the illustration, the system selects a goal (an image) and encodes it into a lower-dimensional features vector using the convolutional autoencoder. The compressed goal is fed into the inverse model, which infers the required motor command to reach it. An efferent copy of the motor command is fed into the forward model, which estimates the sensory input that would be observed after the execution of the command. Once the action is performed and the new sensory observation is available, this is compared to the prediction, resulting in a prediction error. Prediction error is thus used to update the learning progress of the current goal. The system thus decides whether to keep exploring this goal or to switch to another, according to their expected learning progress. The sensorimotor data gathered throughout the exploration behaviour are used to update the inverse and forward models, and the episodic memory, in an online fashion. See section \ref{sec:algorithm} for further details.}
\label{fig:architecture}
\end{figure*} 

The architecture is illustrated in figure \ref{fig:architecture} and is partially inspired by the intrinsic motivation algorithm presented by \citep{oudeyer2007intrinsic} 
and on our previous works on goal-directed exploration \citep{Schmerling2015}. This paper proposes more advanced computational models compared to those presented by \citep{oudeyer2007intrinsic}, as well as more adaptive exploration strategies that those adopted in \citep{Schmerling2015}.

\subsection{Forward and inverse models}
The architecture is based on two internal models (i.e. forward and inverse models) which encode the dynamics of the sensorimotor system of the robot. Such models are not pre-programmed, but rather trained in an online fashion.

The forward and inverse models are implemented as shallow artificial neural networks that link motor commands to visual inputs and vice versa. The sensory space here is identified as the space of images that can be grabbed from the end-effector camera of the robot. The motor space is identified as the space of the CNC (x,y) motor positions.
We define a goal as a specific state in the sensory space. When the system is given a goal (i.e. a target image), the inverse model is queried to generate the best possible motor commands to reach this goal - i.e. to move the camera to the position where the goal image has been recorded. A copy of this motor command is sent to the forward model. The forward model internally simulates the visual input as if the movement were executed.\footnote{This process is inspired by the classical idea in neuroscience about efference copy and corollary discharge \citep{kawato1999internal,straka2018new,baltieri2018modularity}. We are aware of more recent theories that challenged the usage of efference copies and of inverse models
 \citep{friston2011optimal,feldman2016active,lara2018embodied}. Future works will include reframing the predictive models of this study, getting rid of inverse models and using proprioceptive predictions to generate motor commands, as in the active inference proposal\citep{friston2010free}. As to the current state, we still believe that the proposed approach contributes to the state-of-the-art in intrinsic motivation systems and online learning.
}




Both models are learned and updated in runtime in an incremental fashion. At the beginning of the learning session, the inverse model - whose weights are randomly initialised - is likely to generate random motor commands. This movement produces sensorimotor data that are  used to update the model on the fly. Learning is thus coupled with behaviour, meaning that the training data are produced by the behaviours generated by the models, that are being updated - in parallel - during the activity.

The system calculates a prediction error \textit{PE} by comparing the image predicted by the forward model and the sensory observation captured from the visual input after the execution of the movement. The dynamics of the prediction errors recorded during each attempt to reach a specific goal are monitored. In particular, we use the trend (i.e. the derivative) of the prediction error as an indicator of the expected learning progress:
the larger the \textit{change} in the prediction error, the bigger the expected learning progress. In other words, big changes in the mismatch between what the system expects to perceive and what it observes are interpreted as cues for novelty.


\subsection{Goal selection and learning progress}
\label{sec:goal_selection}
At the beginning of the learning process, the system is given a pre-defined set of goals (i.e. a set of images recorded from different camera positions).
At every iteration, an \textit{interest model} chooses a goal according to a goal selection strategy. This strategy relies on the estimated learning progress that is measured for each goal. That is, the system maintains for each goal an indicator about whether it is advantageous - in terms of information gain - to continue exploring around this goal or not. If exploring a goal is not expected to produce big changes in the prediction error (i.e. the expected learning progress is low), then the interest model drives the exploration towards another interesting goal. 

The expected learning progress $\textrm{LP}$ of a goal $g$, i.e. $\textrm{LP}_g$, is computed as follows:

\begin{equation}
\label{eq:1}
\textrm{LP}_g = \textrm{tanh}(|\textrm{PE}_g(t) - \textrm{PE}_g(t-1)|)
\end{equation}

where $\textrm{PE}_g(t)$ is the prediction error calculated at time $t$. 
At every the iteration, the system choses with a probability of $15\%$ a greedy goal selection strategy, instead of the aforementioned one (see \citep{oudeyer2007intrinsic} for the rationale behind this approach, and \citep{tschantz2019learning} for alternative approaches). This strategy selects a random goal from the existing set of goals.

Moreover, with a probability of $30\%$, a random movement is generated instead of the predicted camera movement. This is performed to prevent the system from converging towards local minima 
\footnote{This may not be optimal, especially when the motor space is high-dimensional. See \citep{tschantz2019learning} for more details.}.


\subsection{Image encoding}
\label{sec:goals}

Using images as sensory states raises the issue about how to measure dissimilarity between predictions and observations. Images are high-dimensional data. 
Typically, in machine learning, dissimilarity between images is not computed by pixelwise comparisons, but rather by comparing features extracted from them \citep{chen2005similarity}.
In this work, we used an unsupervised learning technique to extract low-dimensional features from images, so that the resulting features can be simply compared using Euclidean distances. We adopted deep convolutional autoencoders (CAEs) for this task. 
A convolutional autoencoder \citep{masci2011stacked} is a deep neural network that is trained to reproduce the same data passed as input. The dimensionality of the layers in the network decreases from the input layer to a central one (the \textit{latent layer}), and then increases back to the original dimensionality at the output layer. This enables 
 the learning of low-dimensional features in the latent layer, representing the input data. Autoencoders are typically used for compressing high-dimensional data, such as raw images, into low-dimensional codes, and eventually for reconstructing the input from these features.

Autoencoders generally have fully-connected layers. Convolutional autoencoders extend the original structure of autoencoders by using convolutional layers. In convolutional layers, the connectivity between neurons resembles the organisation of the human visual cortex, where neurons respond only to limited parts of the visual field, known as the \textit{receptive field}. Beside being more biologically plausible, convolutional layers considerably reduce the number of parameters that need to be learned by the artificial neural network and have been demonstrated to perform much better in computer vision applications. 
%

The usage of the CAE allows a simpler calculation of prediction errors. Calculating prediction errors corresponds to computing the Euclidean distance between the compressed representations of the predicted image and the observed one.

\subsection{Episodic memory}
\label{sec:epi_mem}
As mentioned in the introduction of this paper, training artificial neural networks in an online fashion may produce catastrophic forgetting issues. In the current architecture, the shallow neural networks used in the forward and inverse model are trained on new tasks on the fly.
Such training on new samples may disrupt connection weights that were encoding previous mappings \citep{Masse2018}. Different strategies have been proposed to overcome this problem. We adopt an episodic memory system that maintains a subset of previously experienced sensorimotor data and replays them, along with the new samples, to the networks during the training. This approach is known as memory consolidation or system-level consolidation \citep{mcclelland1995there}. 

The episodic memory is fed with samples observed during the behaviour of the artificial agent. In particular, the episodic memory is empty at the beginning of the learning session. New samples, as observed from the camera and from the motor system (both sensorimotor information are stored into a memory element), are added into the memory, as soon as this reaches its full size (as described in Section \ref{sec:experiments}, we vary this parameter in our experiments). When the memory is full, the system keeps adding new samples into it, whenever available, discarding old elements of the memory. A greedy strategy is adopted, where random elements of the memory are removed from the list and replaced with new ones. Moreover, a second parameter drives such an episodic memory update: $p_\textrm{em}$, i.e. the probability of changing an element in the episodic memory. The episodic memory update process consists of iterating over all the elements of the memory and, with a probability of $p_\textrm{em}$, of discarding each element and replacing it with the new sample. This produces duplicates of the new samples in the memory, which may increase the plasticity of the system towards latest observations. 
 
In Sections \ref{sec:experiments} and \ref{sec:results}, we show the impact of different configurations of the episodic memory to the whole learning process, and to the behaviour of the simulated robot.

\subsection{The learning algorithm}
\label{sec:algorithm}
Using the components described above, we can now describe the basic learning algorithm as follows:

\begin{enumerate}

\item The system selects $N$ goals, $\textrm{g}_i$, following the strategy described in section \ref{sec:goal_selection}. Each goal is the compressed representation of an image produced by the pre-trained convolutional autoencoder (CAE). The prediction errors $\textrm{PE}(\textrm{g}_i,0)$ are initialized to 0, and $\textrm{LP}(\textrm{g}_i,0)$ to 0, for each goal. 

\item The interest model selects the goal $\textrm{g}(t)$ in $\{\textrm{g}_i\}$ which has the maximum expected learning progress $\textrm{LP}(g_i, t)$. The first goal $\textrm{g}(0)$ of the learning process is randomly selected from $\{\textrm{g}_i\}$.


\item The inverse model computes the new position of the camera $\textrm{P}(t)$ with the objective that the (encoded) image of the camera after the movement corresponds to the selected goal.

\item The position $\textrm{P}(t)$ is sent to the forward model. The forward model predicts the compressed image $\textrm{I}_p(t)$ that will be seen at the new position.

\item The camera is moved. A random noise is added to the camera movement predicted by the inverse model, in order to produce small fluctuations and thus to generate more views of the target locations.

\item The camera image $\textrm{I}_c(t)$ is grabbed and compressed by the CAE.

\item The prediction error $\textrm{PE}(t)$ is computed as the Euclidean distance between $\textrm{I}_p(t)$ and $\textrm{I}_c(t)$ (the compressed representations of the predicted and current images).

\item The estimated learning progress is updated using equation~\ref{eq:1} and $\textrm{PE}(t)$.

\item The inverse model is updated with the recorded sensorimotor data and the elements stored in the memory.

\item The forward model is updated with the recorded sensorimotor data and the elements stored in the memory.

\item Update the episodic memory, as described in section \ref{sec:epi_mem}.

\item Return to step 1, until the maximum number of iterations is reached (5000 in the current experiment).
\end{enumerate}

When the camera moves to a new position computed by the inverse model, the difference between the expected image and the obtained image may be very large. The prediction error $\textrm{PE}$ will be large in that case. If the previous action of reaching the same goal produced a prediction error of different magnitude, this results in a high estimated learning progress $\textrm{LP}$: the system's guess about the consequences of the current action is not met, and it also changes very much when re-iterating it, suggesting that such an action is very informative.
The interest model is then likely to select the same goal for the next iteration (step 2). 
In case the current activity does not produce big changes in the prediction error, compared to the previous ones, the interest model selects another goal, as the current one does not bring any novel information to the learning process.

It has to be noted that we do not update the inverse and forward models at each iteration but only after a given batch size has been completed. The fitting is then done using the history of sensory and motor data for that batch. We use batches for performance reasons and for preventing less accurate gradient estimations due to the big fluctuations that would be produced, otherwise, by stochastic learning (i.e. batch size of one sample).

\section{Experiments}
\label{sec:experiments}

As mentioned in the previous sections, we adopt a convolutional autoencoder (CAE) for unsupervised learning of image features. The CAE is pre-trained in an offline fashion, that is ahead of the learning experiment. 
Thereafter, during the online test, image goals are compressed into features, using the encoder part of the pre-trained CAE.

The CAE has been implemented using Keras deep learning library and TensorFlow backend. 
The structure of the CAE is the following.
An input layer takes a $64 \times 64$ grayscale image and passes it to a sequence of $\textrm{conv2D}$ and $\textrm{MaxPooling2D}$ layers: $\textrm{conv2d}_1$ has $64 \times 64 \times 256$ neurons,
$\textrm{conv2d}_2$ has $32 \times 32 \times 128$ neurons,
$\textrm{conv2d}_3$ has $16 \times 16 \times 128$ neurons. The output of $\textrm{conv2d}_3$ is flattened and connected to a $\textrm{Dense}$ layer of size 32 (the size of the latent space, i.e. the feature vector representing the compressed image). The decoder part of the CAE starts with connecting the latent later to a $\textrm{Dense}$ layer of size 64, reshaping it into a $16 \times 16$ layer and connecting it to a $\textrm{conv2d}$ layer, characterised by $8 \times 8 \times 32$ neurons. The following sequence of layers is then expanding the latent signal: $\textrm{upsampling2d}$, $\textrm{conv2d}$ ($16 \times 16 \times 128$), $\textrm{upsampling2d}$, $\textrm{conv2d}$ ($32 \times 32 \times 512$), $\textrm{upsampling2d}$, $\textrm{conv2d}$ ($64 \times 64 \times 1$). The latter represents the output \textit{decoded} layer, whose size matches the one of the input.
The $\textrm{MaxPooling}$ kernel has size 2, padding "same". All the $\textrm{conv2D}$ layers have kernel size equals to 3 and ReLU activation functions except the CAE output layer that has a sigmoid activation function. An ADAM optimiser has been used for training the network on an MSE loss function.

The inverse and forward models have been implemented as shallow artificial neural networks and are trained in an online fashion during the experiment. 
The structure of the forward model is characterised by the following layer. An input layer is fed with a two-dimensional vector (the (x,y) motor command applied to the CNC machine), which is expanded to a $\textrm{dense}$ layer of 32 neurons, followed by two $\textrm{dense}$ layers of 320 neurons, and a $\textrm{dense}$ layer of 32 neurons (the latent code dimensionality), representing the output of the forward model.
The structure of the inverse model is characterised by the following layers. An input layer is fed with a 32-dimensional vector (the latent code, representing the goal image), which is passed to a $\textrm{dense}$ layer of the same size, then expanded to two $\textrm{dense}$ layers of 320 neurons, and compressed back to a $\textrm{dense}$ layer of 2 neurons, representing the output of the inverse model (the (x,y) motor command to apply to the CNC machine). For both the inverse and forward models, the $\textrm{dense}$ layers have $\textrm{tanh}$ activation functions. A standard gradient descent (LR: 0.0014, decay: 0.0, momentum: 0.8) is used as optimiser on an MSE loss function. 
The training of the CAE, the inverse and forward model updates, as well as the predictive processes are run on a machine equipped with an NVIDIA graphics card.


As discussed in the previous sections, we adopt an episodic memory system that maintains a subset of previously experienced sensorimotor data and replays them, along with the new samples, to the networks during the training. As soon as they are available, new samples (image and motor information) are added into the memory. When the memory reaches its full size, random elements are removed from the memory and replaced with new ones (see section \ref{sec:epi_mem} for more details).

The inverse and forward models are updated at constant frequency, that is, every time a new buffer of 16 sensorimotor samples (each consisting of (x,y) motor commands and 32-dimensional compressed image codes gathered along the exploration) is available. In particular, a new fit for both the inverse and forward models is triggered, passing as training data the buffer with the \textit{new} 16 observations together with \textit{all} the samples stored in the episodic memory. 

In the current experiment, we vary two parameters over different runs  and analyse their influence onto the learning process. In particular, we vary the size of the episodic memory and the probability of changing an element in the episodic memory ($p_\textrm{em}$).

\section{Results}
\label{sec:results}
As discussed in the previous section, the experiment consisted of two phases. In an initial offline stage, a convolutional autoencoder has been trained with the images stored in the simulated robot data (24,964 images, see section \ref{sec:simulator}). This neural network has been used to compress the images (sensory inputs and goals) to be used in the online learning process.

\begin{figure}[ht]
	\centering
	\includegraphics[width=8cm]{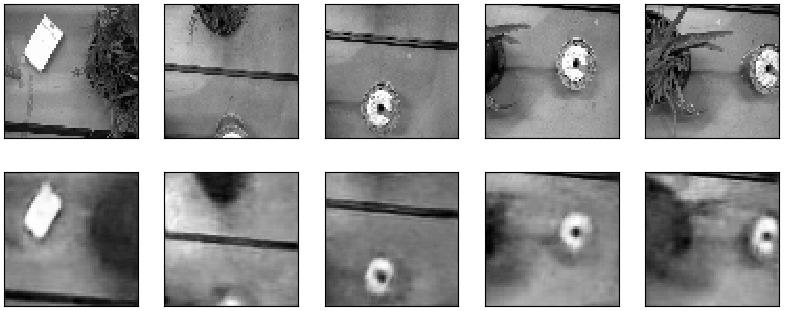}
	\caption{Images encoded and decoded using the convolutional autoencoder. The top row shows the original images. The bottom row shows the result of encoding each input image into a latent code and then decoding it back to the original image resolution.}
	\label{fig:cae_decoded}
\end{figure}

Figure \ref{fig:cae_decoded} shows an example of a sequence of encoded-decoded images produced by such a convolutional autoencoder. As mentioned in the previous section, the image features (i.e. the compressed version of the image) consists of a 32-dimensional vector. Input images and reconstructed images are characterised by a resolution of $64\times64$ pixels. The reconstructed images showed in Figure \ref{fig:cae_decoded} have been generated by a CAE trained for  50 epochs.

Once having trained the CAE, the full learning process can be performed as described in Section \ref{sec:algorithm}. We carried out a series of experiments, in which we varied two parameters of the system: the size of the episodic memory $\textrm{mem}_\textrm{size}$ and the probability of updating the memory elements $p_\textrm{em}$. In particular, we tested three different values for $\textrm{mem}_\textrm{size}: \{0, 10, 20\}$ (with 0 meaning that no episodic memory available) and two values of $p_\textrm{em}: \{0.1, 0.01\}$, for a total of 6 experiments. The values of $\textrm{mem}_\textrm{size}$ specify the number of batches (each batch has 16 sensorimotor samples) that are contained in the memory. Each experiment has been run 5 times, for a total of 30 runs. Each run consisted of 5000 iterations of the algorithm described in Section \ref{sec:algorithm}. During each run, the mean squared error of the forward and inverse models, calculated on the same test dataset (consisting of 50 images randomly chosen from the simulator dataset) have been monitored. The MSEs have been calculated every 50 iterations of the algorithm (behaviour and model updates). A set of 9 goal images have been chosen from the simulator dataset.

\begin{figure*}
	\centering
	\includegraphics[width=5.6cm]{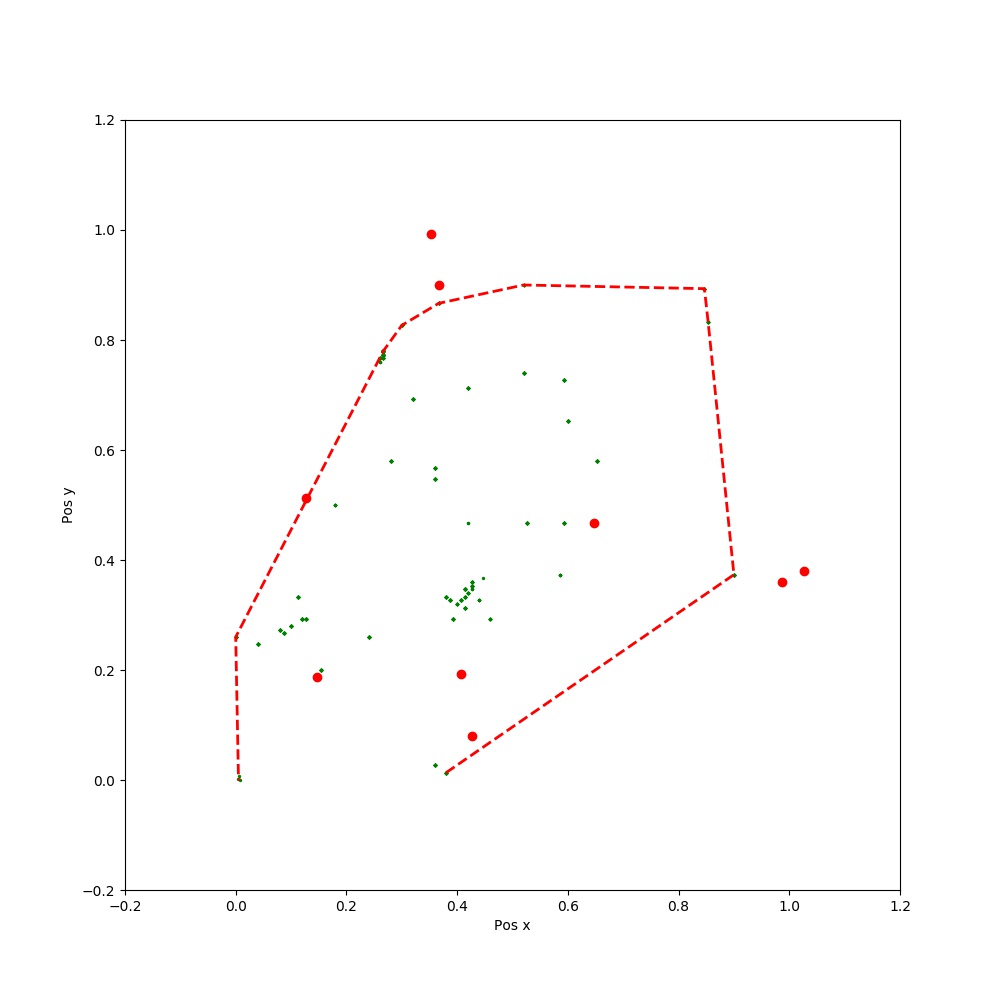}
	\includegraphics[width=5.6cm]{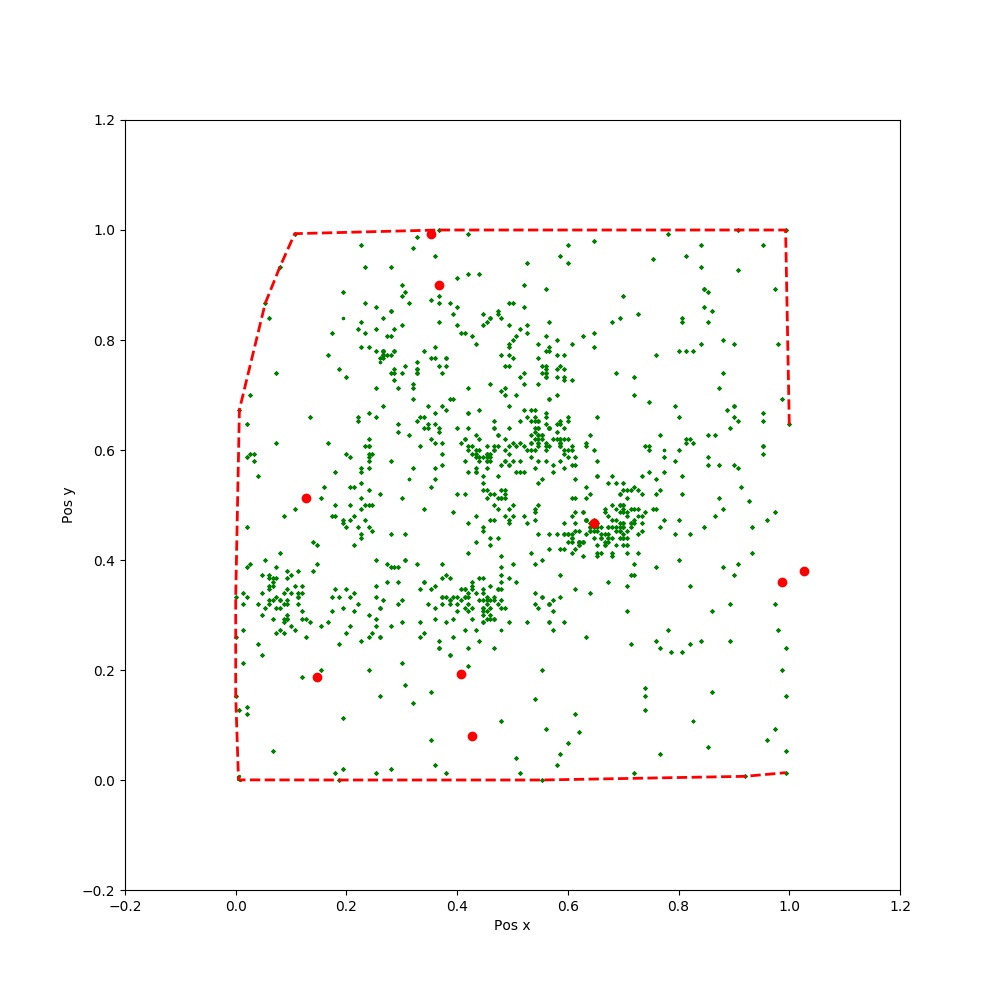}
	\includegraphics[width=5.6cm]{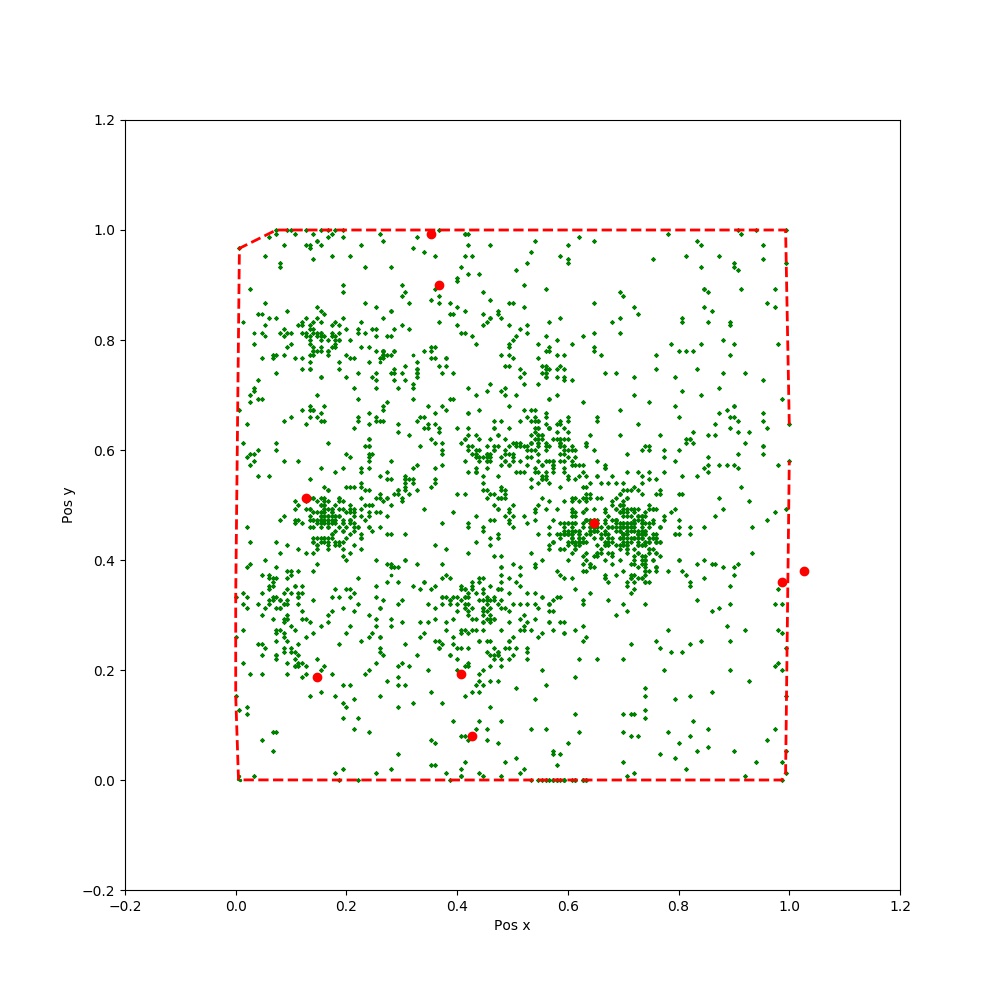}\\
	\includegraphics[width=5.6cm]{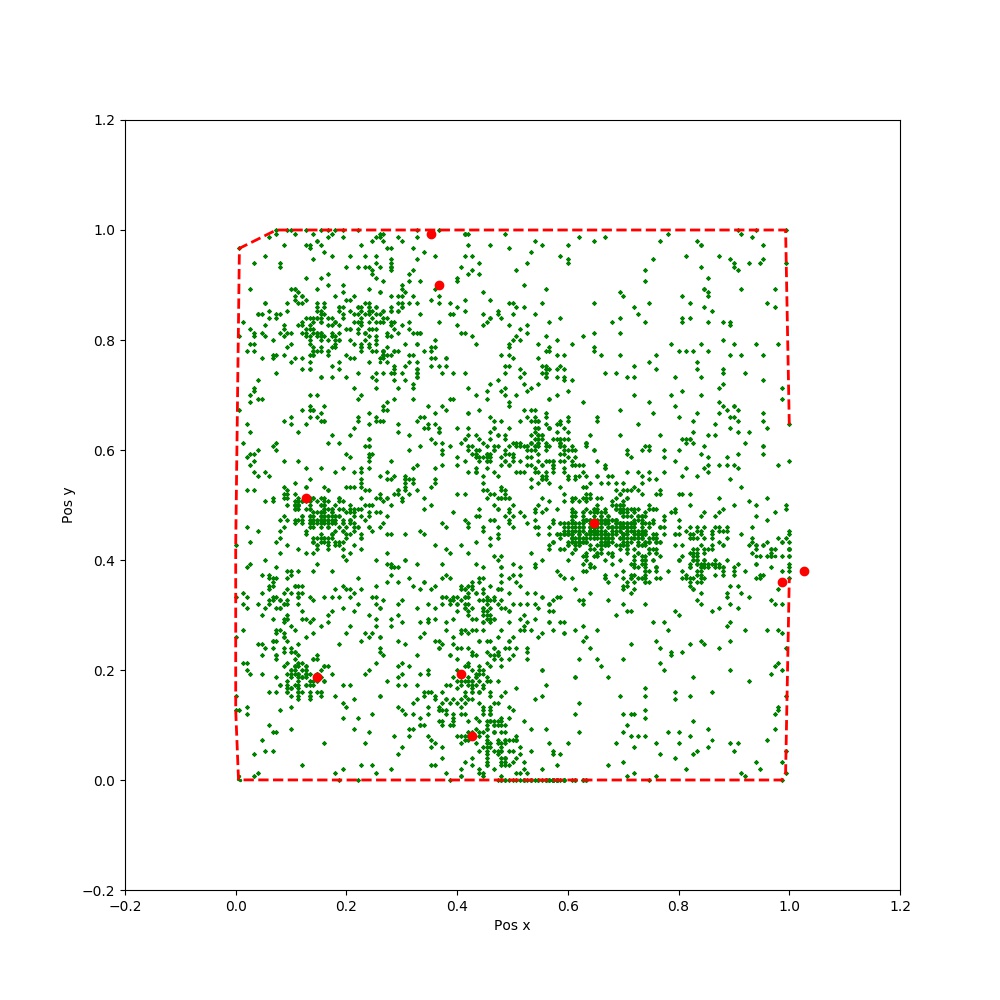}
	\includegraphics[width=5.6cm]{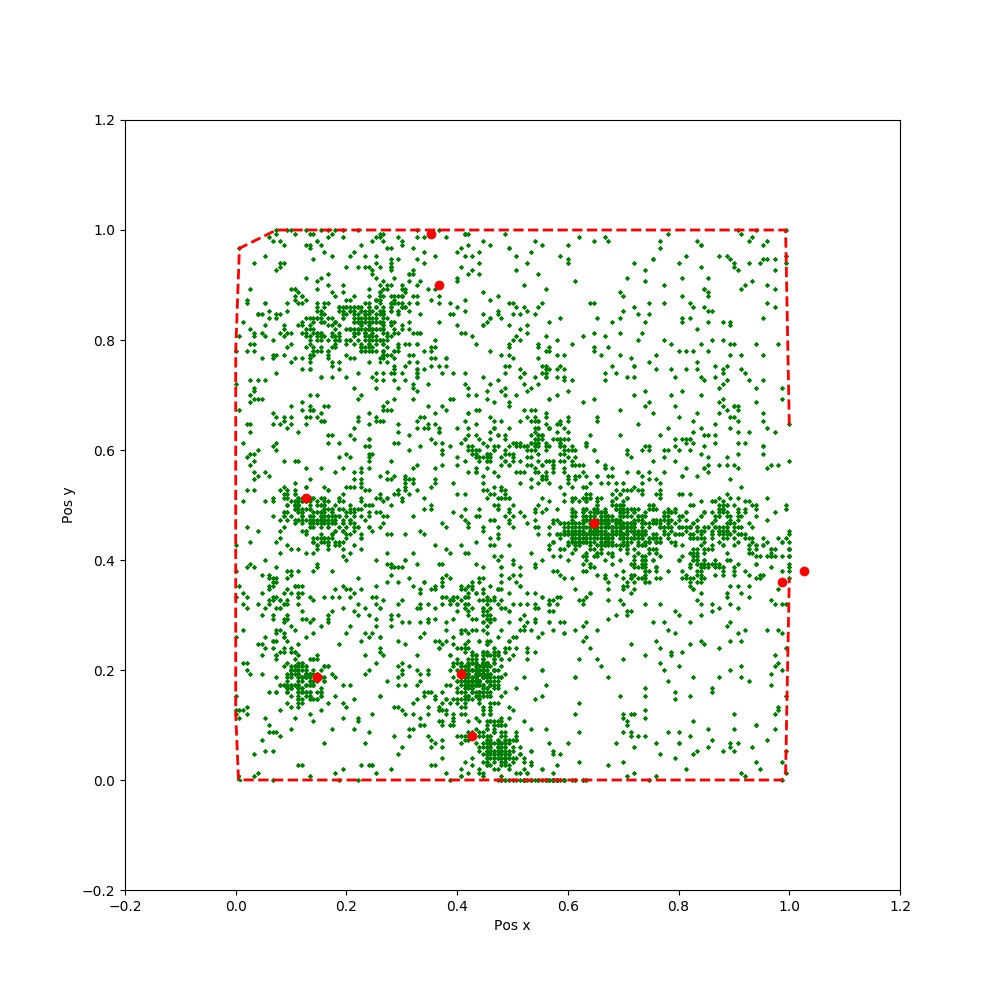}
	\includegraphics[width=5.6cm]{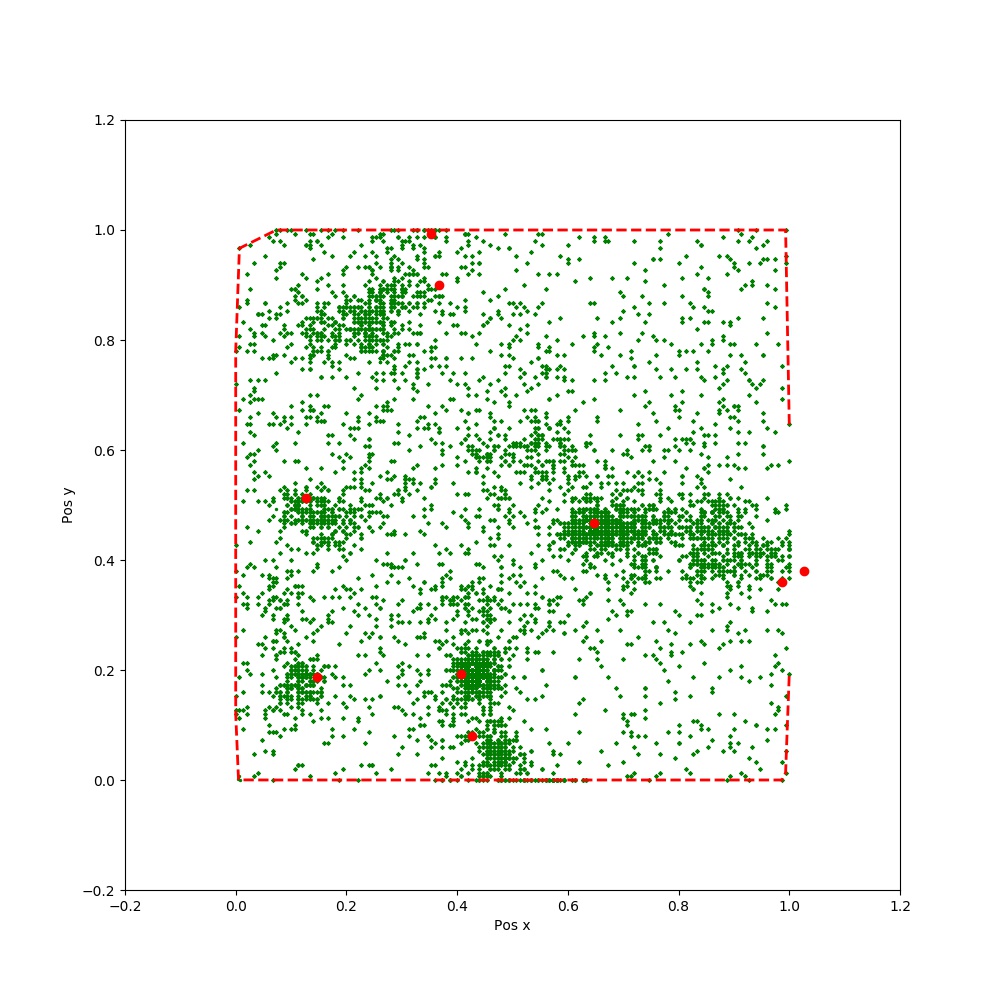}
	\caption{An example of a goal-directed exploration session driven by artificial curiosity (run no. 5, configuration ($\textrm{mem}_\textrm{size}=20$, $p_\textrm{em}=0.1$)). A set of 9 goals are defined from the pre-recorded image dataset. The red dots show the (x,y) motor configurations - which are stored as ground truth data in the dataset - that correspond to the image goals. Each of the plots shows the experienced (x,y) motor commands after 50 (top left plot), 1000, 2000 (top right), 3000 (bottom left), 4000 and 5000 (bottom right) samples. Green dots represent the (x,y) motor positions captured from the CNC machine while exploring. The red dashed line shows the convex hull computed around the explored points.}
	\label{fig:exploration}
\end{figure*}

Figure \ref{fig:exploration} shows an example of a run of the exploration behaviour under a specific configuration ($\textrm{mem}_\textrm{size}=20$, $p_\textrm{em}=0.1$) and over different time steps (50, 1000, 2000, 3000, 4000 and 5000 iterations). The axes specify the x and y motor positions of the CNC machine. Red dots represents the motor projections of the 9 image goals (ground truth data from the simulator dataset). Green dots represent the explored positions of the robot. As it can be seen, the behaviour of the robot successfully tends towards exploring more the regions around the goals. The effect of the $\epsilon$-greedy strategy is also evident from the almost uniformly distributed green points within the action space.

\begin{figure}[ht]
	\centering
	\includegraphics[width=9cm]{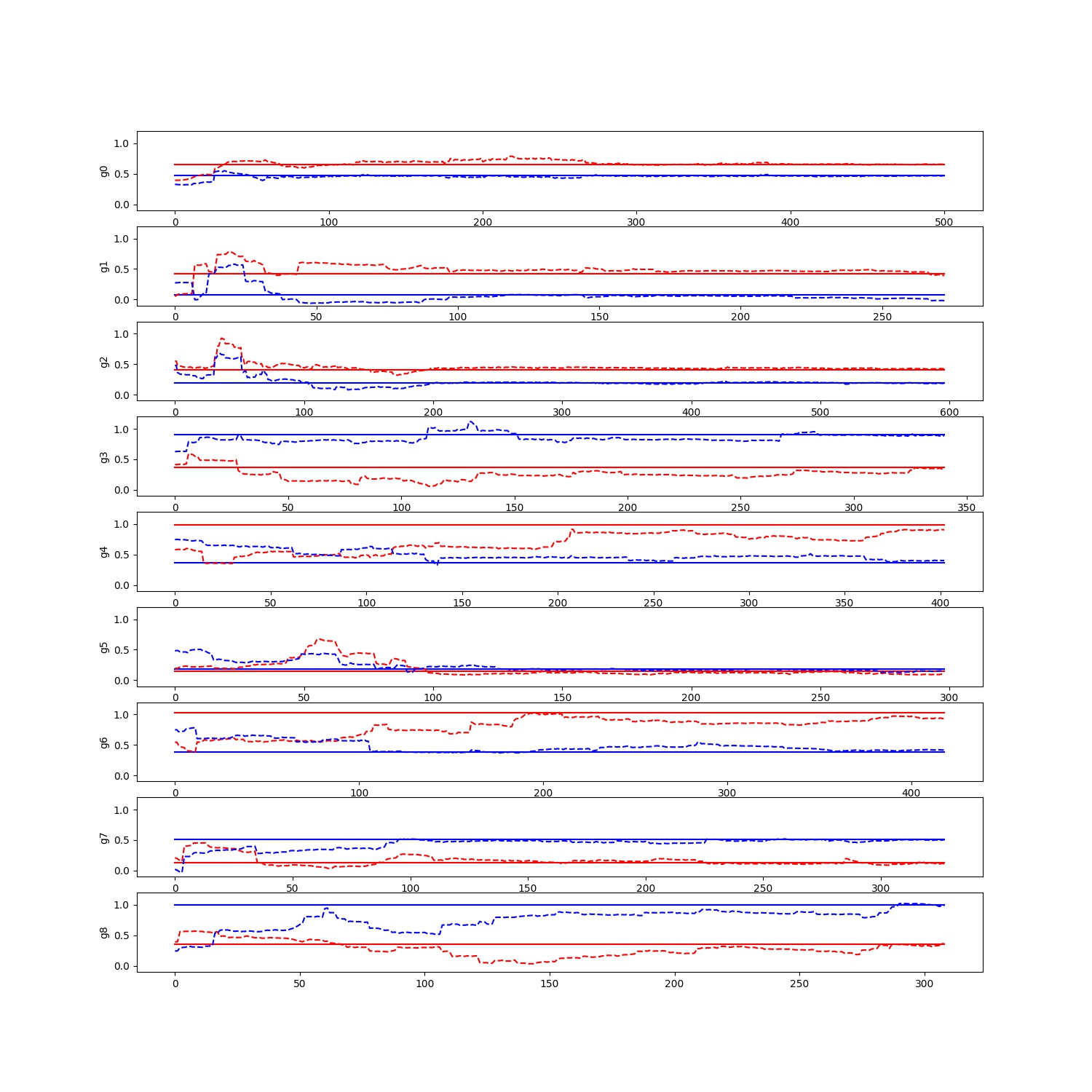}
	\caption{The predictions of the 9 goals over time performed by the inverse model in the run no.5 of the configuration ($\textrm{mem}_\textrm{size}=20$, $p_\textrm{em}=0.1$). The inverse model is fed with the encoded goal image  as input and returns as output the (x: blue,y: red) coordinates of the motor. Solid lines show, for each goal image, the ground truth motor positions. Dashed lines show the predictions of the inverse model.}
	\label{fig:goal_prediction_mem}
\end{figure}

\begin{figure}[ht]
	\centering
	\includegraphics[width=9cm]{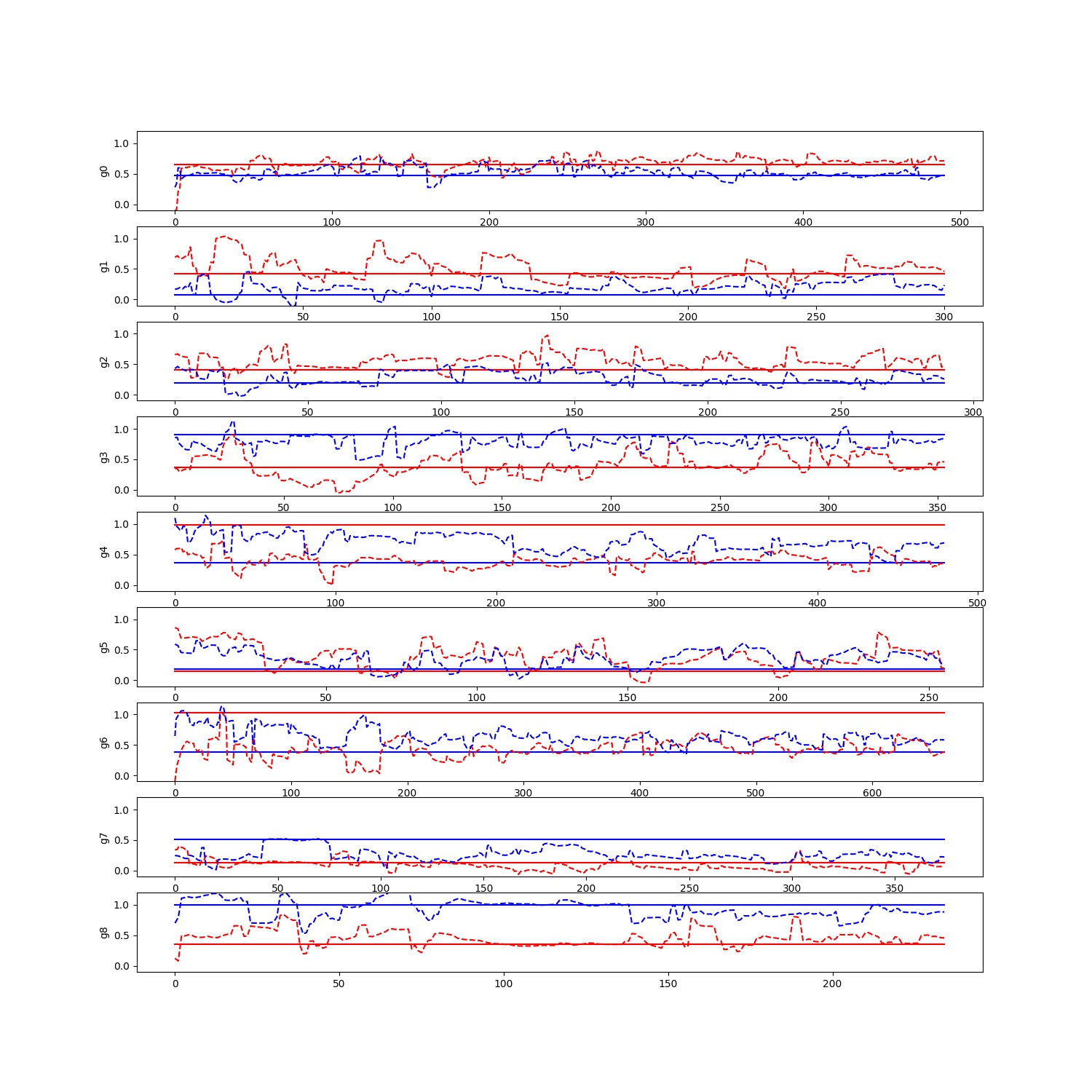}
	\caption{The predictions of the 9 goals over time performed by the inverse model in the run no.0 of the configuration without episodic memory. The inverse model is fed with the encoded goal image  as input and returns as output the (x: blue, y:red) coordinates of the motor. Solid lines show, for each goal image, the ground truth motor positions. Dashed lines show the predictions of the inverse model.}
	\label{fig:goal_prediction_nomem}
\end{figure}

Figure \ref{fig:goal_prediction_mem} shows the predictions of the inverse models over time. The inverse model is fed with the encoded goal image as input and returns as output the (x,y) coordinates of the motor. It can be clearly seen that the learning process brings the prediction towards the target motor goals (ground truth data). The episodic memory makes sure that previous knowledge (when alternating between goals) is not forgotten. The impact of catastrophic forgetting, due to the absence of the episodic memory, is more clear in Figure \ref{fig:goal_prediction_nomem}, which shows the prediction of the inverse model over time. As it can be seen, predictions over different goals are much more noisy than in the previous case. 

\begin{figure}[ht]
	\centering
	\includegraphics[width=9cm]{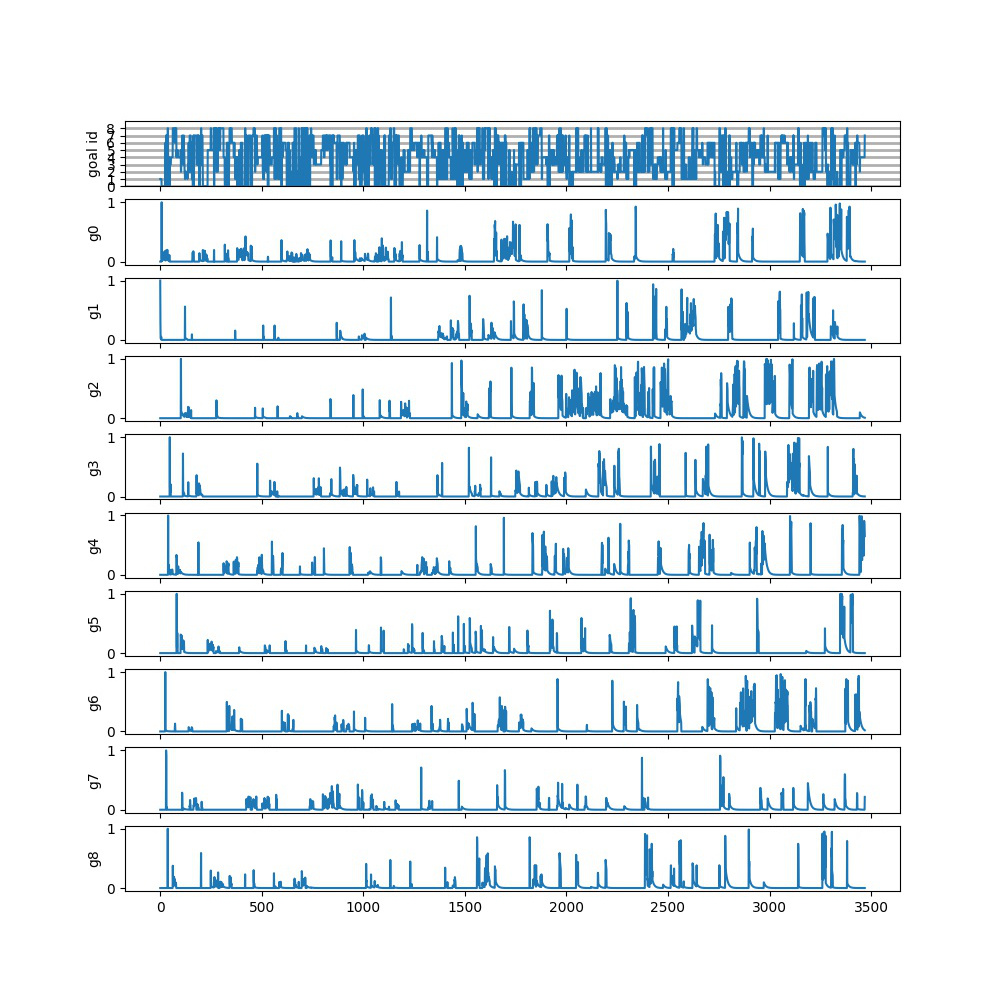}
	\caption{The dynamics of the expected learning progress for each goal (from 2nd to 10th plots). The first plot shows the ID of the currently selected goal over time.}
	\label{fig:lp_dynamics}
\end{figure}

Figure \ref{fig:lp_dynamics} shows the dynamics of the expected learning progress over the 9 goals of the same sample run described above ($\textrm{mem}_\textrm{size}=20$, $p_\textrm{em}=0.1$). Each goal is characterised by the expected learning progress described in equation \ref{eq:1}. A decay factor is also added, so that expected learning progress slowly decays over time. The time scales (horizontal axis) of this plot and those of figures \ref{fig:goal_prediction_mem} and \ref{fig:goal_prediction_nomem} are not matching as predictions are performed only when the specific goal is selected by the intrinsic motivation system. 

\begin{figure}[ht]
	\centering
	\includegraphics[width=8cm]{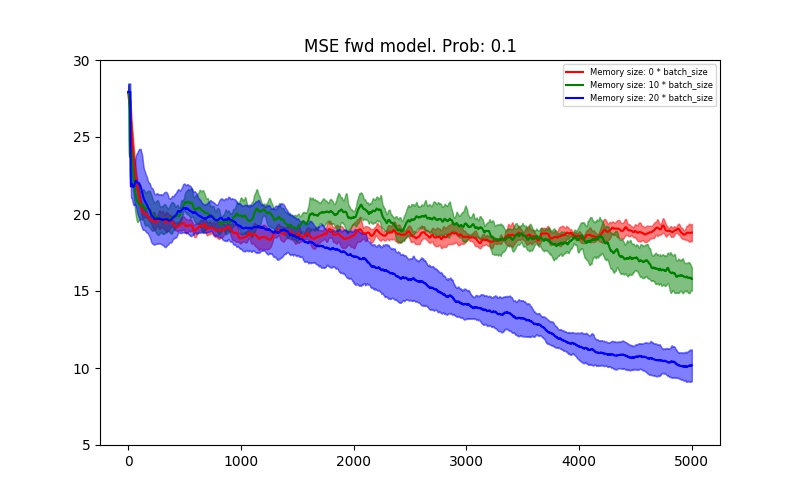}
	\caption{The mean squared error of the forward model computed over 5000 iterations, averaged on 5 runs. The plot compares the MSE of three configurations of the system: without memory (0 batches), $\textrm{mem}_\textrm{size}=10$ and $\textrm{mem}_\textrm{size}=20$. In this plot, the probability of changing memory elements is set to $p_\textrm{em}=0.1$. Solid lines show the mean of the MSE over 5 runs, for each configuration. Shaded areas show the ($\textrm{mean}-\textrm{stddev}$, $\textrm{mean}+\textrm{stddev}$) areas.}
	\label{fig:mse_fwd_0.1}
\end{figure}

\begin{figure}[ht]
	\centering
	\includegraphics[width=8cm]{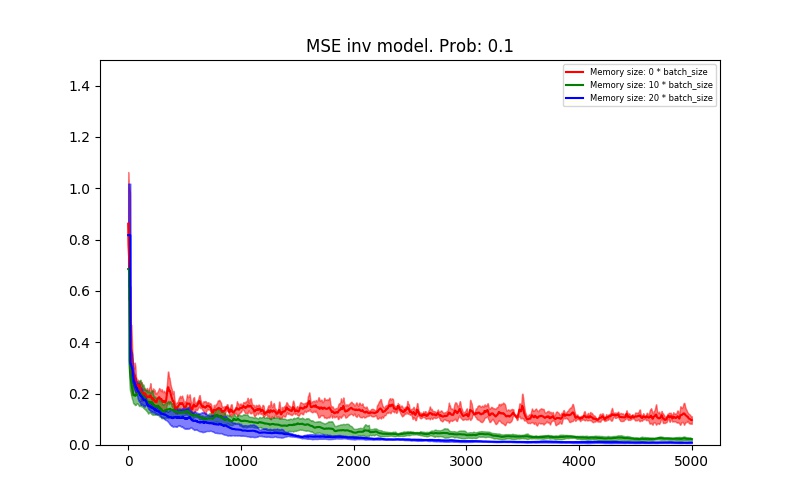}
	\caption{The mean squared error of the inverse model computed over 5000 iterations, averaged on 5 runs. The plot compares the MSE of three configurations of the system: without memory (0 batches), $\textrm{mem}_\textrm{size}=10$ and $\textrm{mem}_\textrm{size}=20$. In this plot, the probability of changing memory elements is set to $p_\textrm{em}=0.1$. Solid lines show the mean of the MSE over 5 runs, for each configuration. Shaded areas show the ($\textrm{mean}-\textrm{stddev}$, $\textrm{mean}+\textrm{stddev}$) areas.}
	\label{fig:mse_inv_0.1}
\end{figure}

Figure \ref{fig:mse_fwd_0.1} compares the learning progress of the forward model in three different configurations of the episodic memory. As it can be seen, the intrinsic motivation system does not work well without episodic memory (red curve). Adding an episodic memory has a positive impact on the learning progress, as the mean squared error of the forward model predictions decreases over time (bigger memory generates better results). Similar trends can be observed for the inverse model (Figure \ref{fig:mse_inv_0.1}).

\begin{figure}[ht]
	\centering
	\includegraphics[width=8cm]{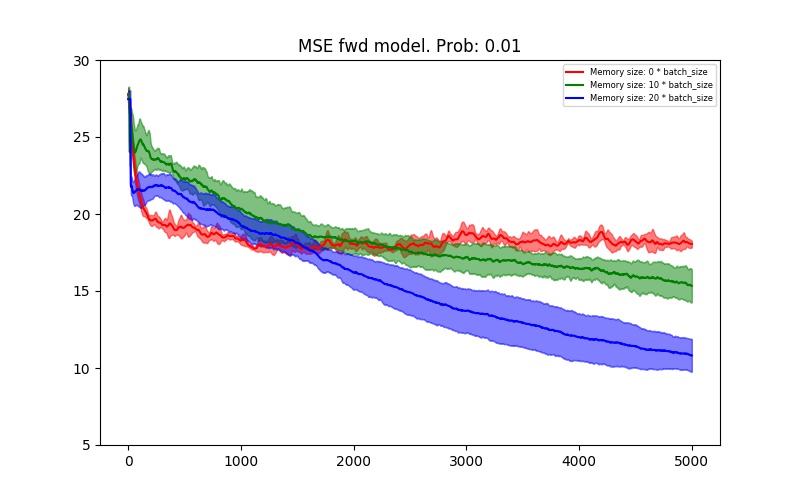}
	\caption{The mean squared error of the forward model computed over 5000 iterations, averaged on 5 runs. The plot compares the MSE of three configurations of the system: without memory (0 batches), $\textrm{mem}_\textrm{size}=10$ and $\textrm{mem}_\textrm{size}=20$. In this plot, the probability of changing memory elements is set to $p_\textrm{em}=0.01$. Solid lines show the mean of the MSE over 5 runs, for each configuration. Shaded areas show the ($\textrm{mean}-\textrm{stddev}$, $\textrm{mean}+\textrm{stddev}$) areas.}
	\label{fig:mse_fwd_0.01}
\end{figure}

\begin{figure}[ht]
	\centering
	\includegraphics[width=8cm]{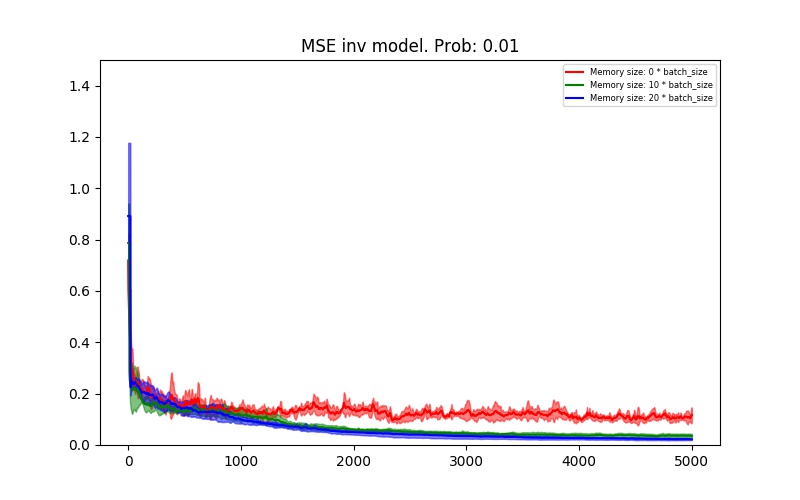}
	\caption{The mean squared error of the inverse model computed over 5000 iterations, averaged on 5 runs. The plot compares the MSE of three configurations of the system: without memory (0 batches), $\textrm{mem}_\textrm{size}=10$ and $\textrm{mem}_\textrm{size}=20$. In this plot, the probability of changing memory elements is set to $p_\textrm{em}=0.01$. Solid lines show the mean of the MSE over 5 runs, for each configuration. Shaded areas show the ($\textrm{mean}-\textrm{stddev}$, $\textrm{mean}+\textrm{stddev}$) areas.}
	\label{fig:mse_inv_0.01}
\end{figure}

Reducing the probability of changing memory elements to $p_\textrm{em}=0.01$ produces smoother descending trends in the MSE of the configurations with episodic memory, for both the forward (Figure \ref{fig:mse_fwd_0.01}) and the inverse (Figure \ref{fig:mse_inv_0.01}) models. We explain this effect as a result of a smaller level of plasticity in the episodic memory. As explained in the previous section, the current episodic memory update strategy produces duplicates of the new observations in the memory. Higher values of $p_\textrm{em}$ produces more duplicates of the same samples, and may reduce the variance within the episodic memory.

\begin{figure}[ht]
	\centering
	\includegraphics[width=8cm]{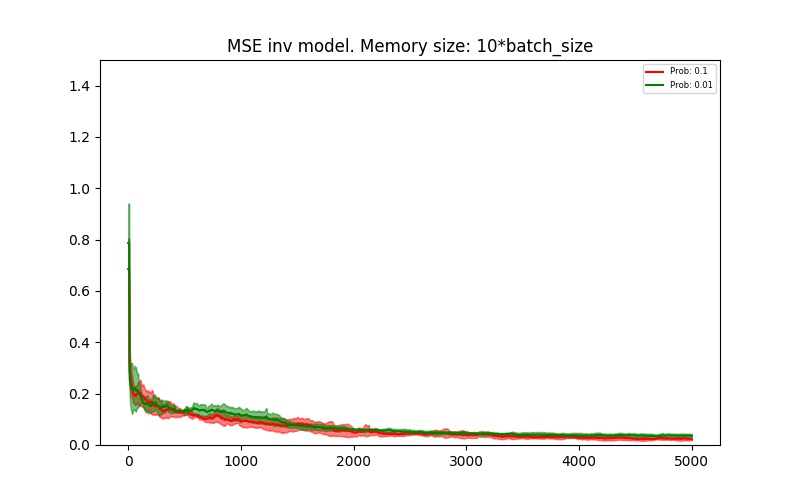}
	\caption{The mean squared error of the inverse model computed over 5000 iterations, averaged on 5 runs. The plot compares the MSE of two configurations of the system: $p_\textrm{em}=0.1$ and $p_\textrm{em}=0.01$. In this plot, the episodic memory size is set to 10 batches. Solid lines show the mean of the MSE over 5 runs, for each configuration. Shaded areas show the ($\textrm{mean}-\textrm{stddev}$, $\textrm{mean}+\textrm{stddev}$) areas.}
	\label{fig:mse_mem_inv_10}
\end{figure}

\begin{figure}[ht]
	\centering
	\includegraphics[width=8cm]{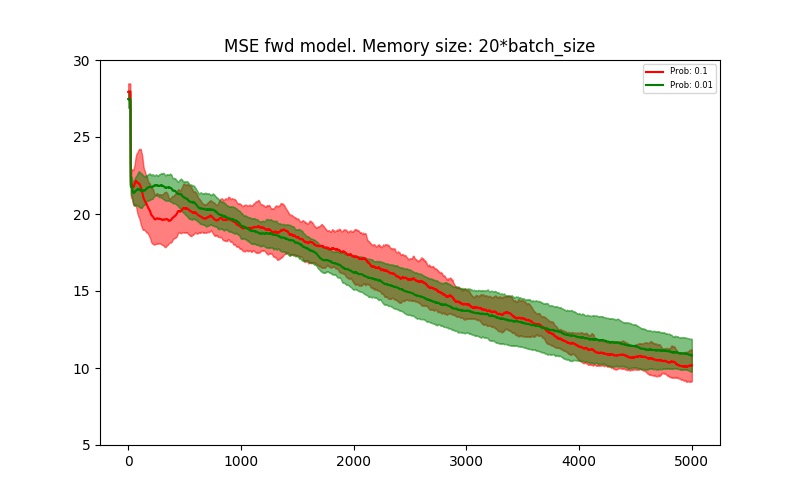}
	\caption{The mean squared error of the forward model computed over 5000 iterations, averaged on 5 runs. The plot compares the MSE of two configurations of the system: $p_\textrm{em}=0.1$ and $p_\textrm{em}=0.01$. In this plot, the episodic memory size is set to 20 batches. Solid lines show the mean of the MSE over 5 runs, for each configuration. Shaded areas show the ($\textrm{mean}-\textrm{stddev}$, $\textrm{mean}+\textrm{stddev}$) areas.}
	\label{fig:mse_mem_fwd_20}
\end{figure}

\begin{figure}[ht]
	\centering
	\includegraphics[width=8cm]{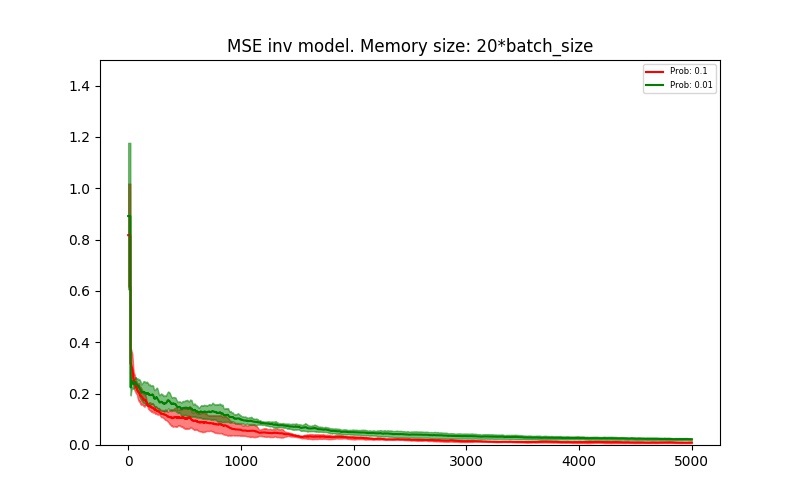}
	\caption{The mean squared error of the inverse model computed over 5000 iterations, averaged on 5 runs. The plot compares the MSE of two configurations of the system: $p_\textrm{em}=0.1$ and $p_\textrm{em}=0.01$. In this plot, the episodic memory size is set to 20 batches. Solid lines show the mean of the MSE over 5 runs, for each configuration. Shaded areas show the ($\textrm{mean}-\textrm{stddev}$, $\textrm{mean}+\textrm{stddev}$) areas.}
	\label{fig:mse_mem_inv_20}
\end{figure}

This can be also seen when comparing the MSEs of the forward and inverse model by pivoting on the $p_\textrm{em}$ values, as shown in the following figures. In Figure \ref{fig:mse_mem_fwd_10}, for instance, it can be seen that the system with higher probability of memory updates (red curve) has a steeper descending curve, although on the long run is outperformed by the other strategy (both the configuration are characterised by a memory size of 10 batches). This suggests that updating the episodic memory more quickly produces higher plasticity, whereas doing it more slowly produces more stability. This effect is however not clearly visible in the MSE curves of the inverse model (Figure \ref{fig:mse_mem_inv_10}). Increasing the memory size to 20 batches leverages this different effect in the forward model, as it can be seen from Figure \ref{fig:mse_mem_fwd_20}, but makes it slightly more evident in the inverse model MSE trends (Figure \ref{fig:mse_mem_inv_20}).

\begin{figure}[ht]
	\centering
	\includegraphics[width=8cm]{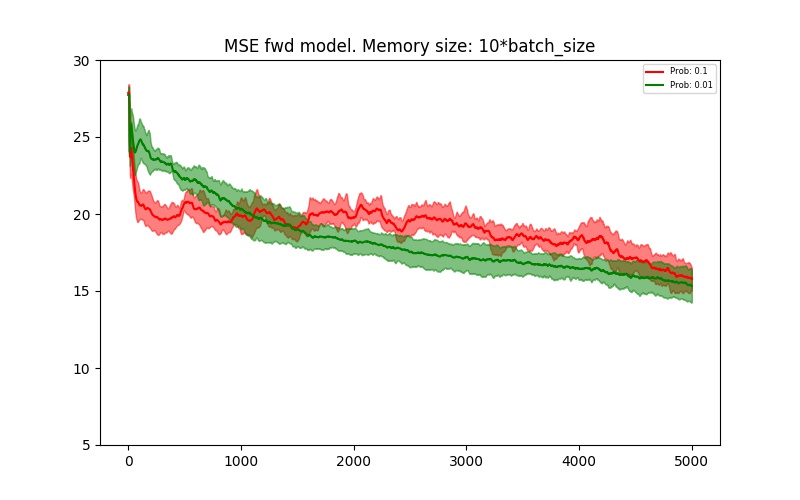}
	\caption{The mean squared error of the forward model computed over 5000 iterations, averaged on 5 runs. The plot compares the MSE of two configurations of the system: $p_\textrm{em}=0.1$ and $p_\textrm{em}=0.01$. In this plot, the episodic memory size is set to 10 batches. Solid lines show the mean of the MSE over 5 runs, for each configuration. Shaded areas show the ($\textrm{mean}-\textrm{stddev}$, $\textrm{mean}+\textrm{stddev}$) areas.}
	\label{fig:mse_mem_fwd_10}
\end{figure}

Further studies will focus on a deeper analysis of the variance of the elements in the episodic memory, 
as well as on more sophisticated strategies for memory update and their effect on the plasticity and stability of the models.

\section{Conclusions}
\label{sec:conclusions}
We presented a learning architecture that generates curiosity-driven goal-directed behaviours based on intrinsic motivation and on an episodic memory system. 
In particular, we implemented online learning mechanisms on artificial neural networks using an episodic memory system for facing catastrophic forgetting issues. We adopted and trained a convolutional autoencoder for compressing image goals and observations into low-dimensional features, that can be more easily processed by our models. We showed the performance of our system under different configurations, where plasticity and stability of the learning process can be modulated. 
Of fundamental importance were the predictive processes implemented through the forward models. In fact, processes of anticipation of sensorimotor activity have enabled the advanced behaviours showed in our experiments. We strongly believe that similar mechanisms may represent a bridge between motor and cognitive development in humans, and a promising tool for cognitive robotics.
We tested the models on simulated sensorimotor data from a microfarming robot. 
This paper has contributed also to the widening of the developmental robotics approach towards applications and robotic platforms that are non-conventional in this field.

Moreover, our experiments have shown that the intrinsic motivation systems can work also in the presence of a high-dimensional sensory space. Adopting an episodic memory system not only prevents the computational models to quickly forget knowledge that has been previously acquired, but also provides new avenues for modulating the balance between plasticity and stability of the system.

Several potential research directions are open from this study. First, the adoption of alternative predictive processes. As discussed in the introduction, more recent approaches have been proposed in the literature, which would better leverage instrumental, goal-directed actions with epistemic, novelty-seeking, behaviour (i.e. active inference \citep{friston2010free,tschantz2019learning}). This proposal has not been yet implemented into high-dimensional sensory spaces and robots such as the ones addressed in this study. 

Second, to achieve the online learning process, image features should be learned in an online and incremental fashion. Similarly, goals could be autonomously generated through unsupervised learning techniques. We are currently exploring different possibilities, including the usage of self-organising maps, trained in an online fashion on the latent codes of the convolutional autoencoder. Goals would be aligned to the -moving- neurons of the self-organising map. Interesting insights could emerge from applying intrinsic motivation systems on dynamic goals, that are autonomously generated by the learning process. 
More advanced goal selection strategies could be implemented, where prediction error dynamics could be analysed over longer time lapses.

Third, similar investigations should be carried out on a multimodal level. The literature on intrinsic motivation systems and goal-directed behaviours in robotics is mostly, if not totally, focusing on unimodal sensory spaces. Investigating these algorithms in the context of multimodal sensory spaces could open very interesting research directions, for instance on how to leverage the competition between modalities in the estimation of the learning progress or in the definition and selection of goals.

\begin{acks}
This work has partially received funding from the European Union's Horizon 2020 research and innovation programme under grant agreement No 773875 (EU-H2020 ROMI, Robotics for Microfarms).

Guido Schillaci has received funding from the European Union’s Horizon 2020 research and innovation programme under the Marie Sklodowska-Curie grant agreement No. 838861 (Predictive Robots).

The authors would like to thank Bruno Lara and Alejandra Ciria for the very helpful feedback on the manuscript.

\textbf{Author contributions}: GS has conceived the presented ideas, implemented them and written them into the manuscript. APV has contributed to the development of the robot simulator. VVH has contributed to the development of the ideas and to the revision of the manuscript. PH, DC and TW have developed the LettuceThink robotic platform and contributed to the revision of the manuscript.

\end{acks}

\bibliographystyle{SageH}
\bibliography{biblio.bib}

\end{document}